\documentclass[a4paper,UKenglish,cleveref, autoref, thm-restate]{lipics-v2019}
\usepackage{url}
\usepackage{amsmath}
\usepackage{amssymb}
\usepackage{amsthm}
\usepackage{color}
\usepackage{array}
\usepackage{xy}
\renewcommand{\paragraph}[1]{\vspace{.5 cm} \noindent \textbf{#1.} }
\newcommand{\defn}[1]{\textbf{\emph{#1}}}

\usepackage{setspace}
\usepackage{multicol}
\usepackage{hyperref}
\usepackage{cite}
\newcommand{\E}{\mathbb{E}}

\newtheoremstyle{slanted}
{3pt}
{3pt}
{\slshape}
{}
{\bfseries}
{.}
{.5em}
{}
\theoremstyle{slanted}
\newtheorem{thm}{Theorem}[section]

\theoremstyle{remark}

\usepackage{fancyhdr}
\begin{document}


\title{Train Tracks with Gaps: Applying the Probabilistic Method to Trains}
\author{William Kuszmaul}{MIT CSAIL, USA}{kuszmaul@mit.edu}{}{}

\funding{Funded by a Fannie and John Hertz Fellowship and an NSF GRFP fellowship. Research also was sponsored by the United States Air Force Research Laboratory and was accomplished under Cooperative Agreement Number FA8750-19-2-1000. The views and conclusions contained in this document are those of the authors and should not be interpreted as representing the official policies, either expressed or implied, of the United States Air Force or the U.S. Government. The U.S. Government is authorized to reproduce and distribute reprints for Government purposes notwithstanding any copyright notation herein.}

\authorrunning{W. Kuszmaul}

\Copyright{William Kuszmaul}



\category{} 

\relatedversion{} 

\supplement{}

\acknowledgements{The author would like to thank Michael A. Bender,
  Bradley C. Kuszmaul, and Charles E. Leiserson for several useful
  conversations about train tracks. The author would also like to
  thank Jake Hillard for offering his engineering expertise, and
  observing that train tracks with an asymptotically small number of
  pillars would likely encounter practical difficulties in the real
  world.}

\nolinenumbers 


\EventEditors{Martin Farach-Colton, Giuseppe Prencipe, and Ryuhei Uehara}
\EventNoEds{3}
\EventLongTitle{10th International Conference on Fun with Algorithms (FUN 2020)}
\EventShortTitle{FUN 2020}
\EventAcronym{FUN}
\EventYear{2020}
\EventDate{September 28--30, 2020}
\EventLocation{Favignana Island, Sicily, Italy}
\EventLogo{}
\SeriesVolume{157}
\ArticleNo{19}

\maketitle

\begin{abstract}
We identify a tradeoff curve between the number of wheels on a train car, and the amount of track that must be installed in order to ensure that the train car is supported by the track at all times. The goal is to build an elevated track that covers some large distance $\ell$, but that consists primarily of gaps, so that the total amount of feet of train track that is actually installed is only a small fraction of $\ell$. In order so that the train track can support the train at all points, the requirement is that as the train drives across the track, at least one set of wheels from the rear quarter and at least one set of wheels from the front quarter of the train must be touching the track at all times.

We show that, if a train car has $n$ sets of wheels evenly spaced apart in its rear and $n$ sets of wheels evenly spaced apart in its front, then it is possible to build a train track that supports the train car but uses only $\Theta( \ell / n )$ feet of track. We then consider what happens if the wheels on the train car are not evenly spaced (and may even be configured adversarially). We show that for any configuration of the train car, with $n$ wheels in each of the front and rear quarters of the car, it is possible to build a track that supports the car for distance $\ell$ and uses only $O\left(\frac{\ell \log n}{n}\right)$ feet of track. Additionally, we show that there exist configurations of the train car for which this tradeoff curve is asymptotically optimal. Both the upper and lower bounds are achieved via applications of the probabilistic method.

The algorithms and lower bounds in this paper provide simple illustrative examples of many of the core techniques in probabilistic combinatorics and randomized algorithms. These include the probabilistic method with alterations, the use of McDiarmid's inequality within the probabilistic method, the algorithmic Lov\'asz Local Lemma, the min-hash technique, and the method of conditional probabilities.
\end{abstract}

\paragraph{A Gap in the Track} A few years ago, while traveling on a
train, and on only a few hours of sleep, I was staring out the
window. The train crossed a bridge over a road, and the ground was
momentarily replaced by a steep drop. Startled, my sleep-deprived mind
briefly wondered whether there was still a track underneath
us. \emph{Of course there is,} I thought to myself. \emph{Without a
  track, the train car would have fallen into the gap.}

\emph{Ah, no so fast!} responded the latent mathematician inside
me. \emph{If the train car had more than two sets of wheels, then
  perhaps it could cross the bridgeless gap without falling in.} It
was true.

Consider, for example, a train with four sets of wheels: one set in the rear, one set in the front, and one set in each of the first and third quartiles.

\vspace{.5 cm}

\begin{center}
  \includegraphics[scale = .7]{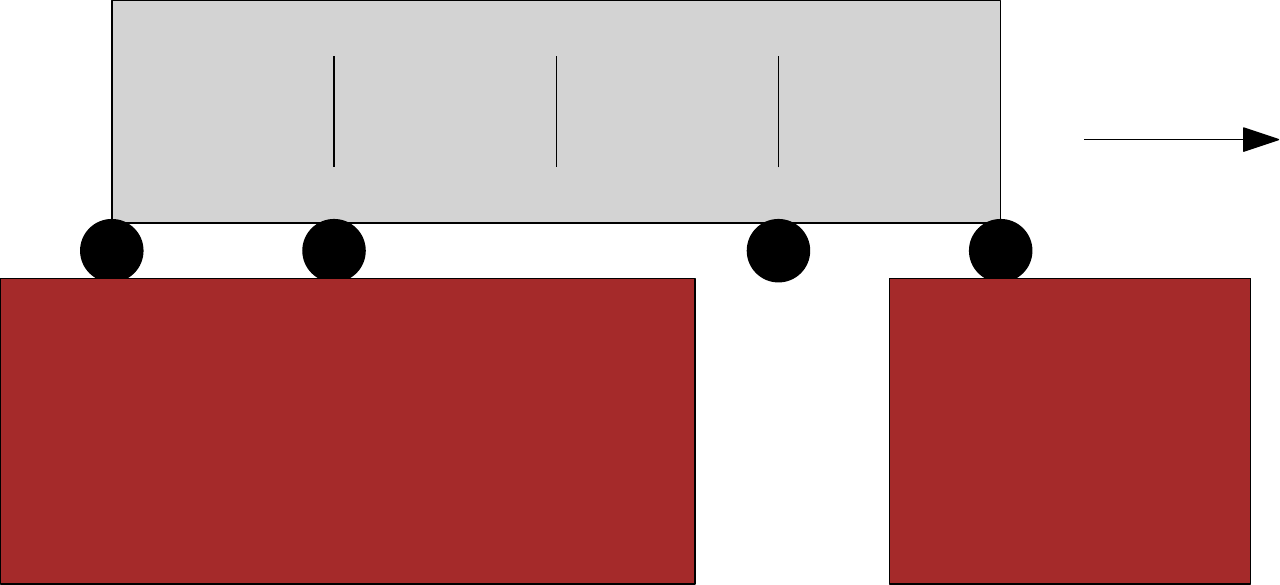}
\end{center}
\vspace{.5 cm}

As long as the gap in the track is less than the distance between any
pair of wheels, then at least three sets of wheels will touch track
at all times. Assuming that the center of mass of the train is in the
middle half of the train, it follows that the train does not fall into
the gap.

\emph{In fact}, continued the latent mathematician, \emph{what if we have $n$ sets of wheels? Maybe we can build a
  mono-rail using an asymptotically small amount of track.}

\emph{That's a stupid idea,} responded I. \emph{Gaps in train track are not something to
  optimize.}

But I was sleep deprived, so I did it anyway.

\paragraph{The Basic Observation: More Wheels Means Less Track} Consider a train car with $2n$
sets of wheels. Half the sets are evenly dispersed across the first
quarter of the train car, and half the sets are evenly dispersed
across the final quarter. The train can safely drive down the track as long as at least one set of wheels from each side of the train car is touching track at all times. The
train car looks something like this:

\vspace{.5 cm}

\begin{center}
  \includegraphics[scale = 1]{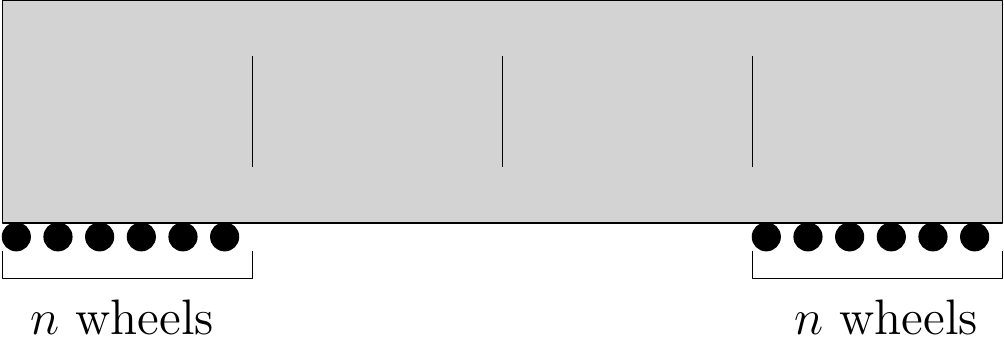}
\end{center}

\vspace{.5 cm}

Want to build a monorail, but you're short on track? No problem! You can get away with filling in only an $O(1/n)$ fraction of the track:

\vspace{.5 cm}

\begin{center}
  \includegraphics[scale = 1]{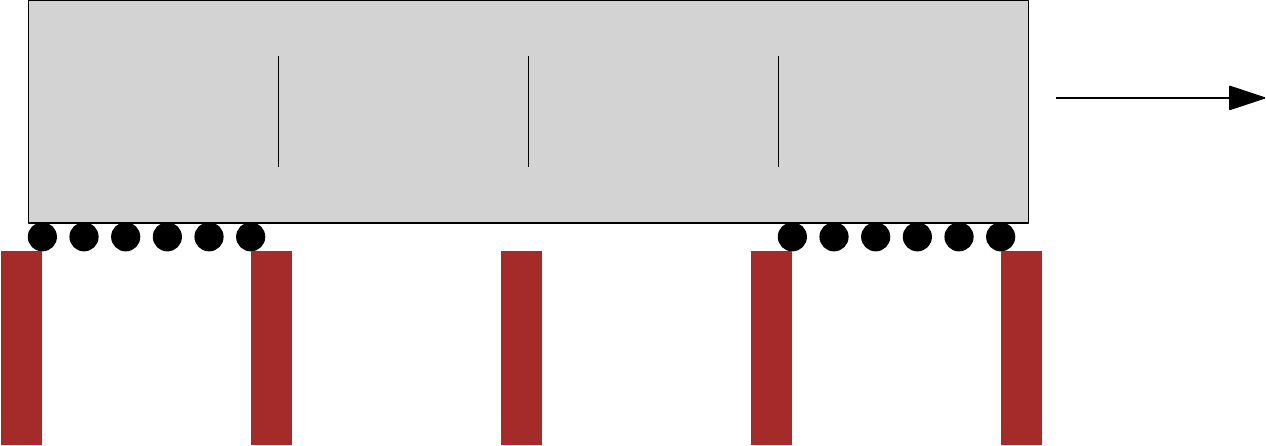}
\end{center}

\vspace{.5 cm}

Every fourth of a train length, we place a piece of track whose length is a $\frac{1}{4n}$ fraction of the length of the train car. We get asymptotic cost savings!

To see that this is the best we can do, suppose that the fraction of track that is filled in is less than $\frac{1}{n}$, and for symmetry sake suppose the track is circular (i.e., the end of the track loops back to the start). If we place the train at a random position in the circular track, then each wheel has a less than $\frac{1}{n}$ chance of touching track. By a union bound, it follows that the probability of any wheel in the rear quarter of the train touching the track is less than $1$. Thus no matter how the track is placed, if the total fraction of track that is filled in is less than $\frac{1}{n}$, then there is some position at which the train falls through.

\paragraph{Paper Outline} The rest of the paper considers the
situation in which the wheels on the train car are placed unevenly
(and possibly even adversarially!) in each of the front and rear
quarters of the car.  Section \ref{sec:upper_bound} describes the problem in more detail, and shows that for any
configuration of the train car, with $n$ wheels in each of the front
and rear quarters of the car, it is possible to efficiently build a
track that supports the car for distance $\ell$ and uses only
$O\left(\frac{\ell \log n}{n}\right)$ feet of track. Section
\ref{sec:lower_bound} then establishes a matching lower bound, showing
that there exist configurations of the train car for which
$\Omega\left(\frac{\ell \log n}{n}\right)$ feet of track are
required. Both the upper and lower bounds are achieved via
applications of the probabilistic method.

The train-track problem serves as a veritable playground for applying
many of the core techniques from probabilistic combinatorics and
randomized algorithms to a simple and fun problem. Section
\ref{sec:three_algorithms} give three alternative algorithms for
achieving the upper bound of $O\left(\frac{\ell \log n}{n} \right)$,
each of which builds on a different technique.

Combined, the algorithms and lower bounds in this paper give simple
illustrative examples of the algorithmic Lov\'asz Local Lemma, the
min-hash technique, the method of conditional probabilities, the
probabilistic method with alterations, and the use of McDiarmid's
inequality within the probabilistic method.

\section{Train Cars with Arbitrary Wheel Arrangements}\label{sec:upper_bound}


Consider a train car that has $n$ wheels in its rear quarter and $n$
wheels in its front quarter, but suppose that the wheels \emph{aren't}
evenly spaced. For example, maybe the rear of the car looks something like this:

\vspace{.5 cm}

\begin{center}
  \includegraphics[scale = 1]{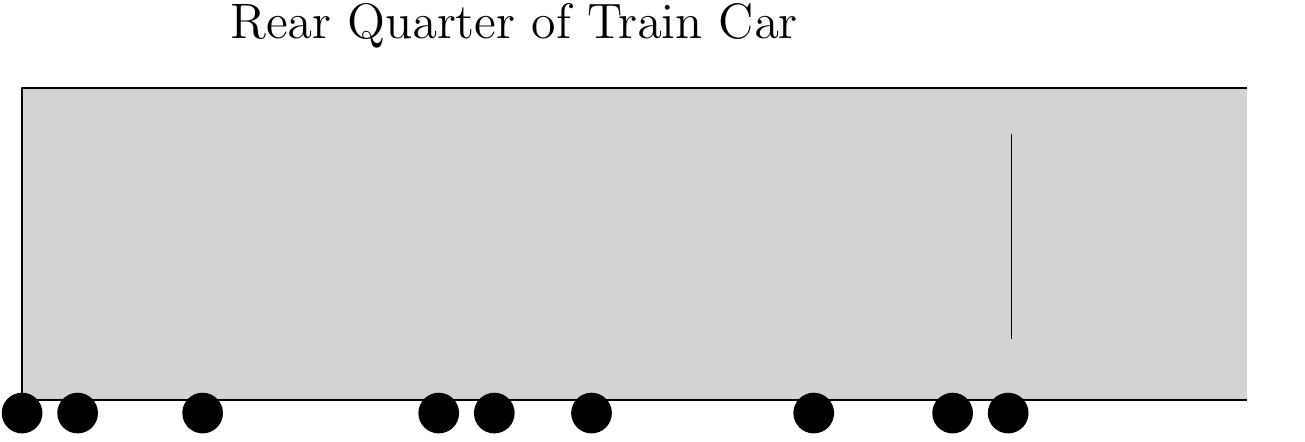}
\end{center}

\vspace{.5 cm}

Can we still fill in an asymptotically small fraction of the track in a way that will allow the train car to drive down the track? In other words, can we place down a small amount of track in a way so that, as the train drives down it there is always at least one wheel from each of the front and rear quarters of the train touching track? It turns out that, via a simple application of the probabilistic method with alterations, we can.

To formalize the situation, let's focus just on the first quarter of the train. (In particular, up to a constant factor in the amount of train track that we install, we can consider the two quarters of the train separately.) Suppose this portion of the train is $f$ feet long, and assume that each of the $n$ sets of wheels resides at a distinct integer offset from the rear of the train. Our goal is to build a train track that is $\ell$ feet long. We build the train track out of \defn{pillars} that are each $1$ foot long and are each placed at integer positions on the track. We are required to put down the pillars in a way so that, as the train drives down the track, at least one wheel from the rear quarter is always touching the track (i.e., touching some pillar). We want to use as little track as possible, with the best we could hope for being a total of $\frac{\ell}{n}$ pillars.

As a reminder, there are three variables: the number of wheels $n$ (in the quarter of the train car that we're considering), the length $f$ of one quarter of the train car, and the length $\ell$ of the train track. In general, we have $n \le f \le \ell$. Note that, although $n$ and $f$ could reasonably be close to one another (e.g., $f = 2n$), we also want to be able to handle cases where $n \ll f$. This allows for the train car to be configured in truly strange ways -- for example, the positions of the wheels could even form a geometric series:

\vspace{.5 cm}

\begin{center}
  \includegraphics[scale = 1]{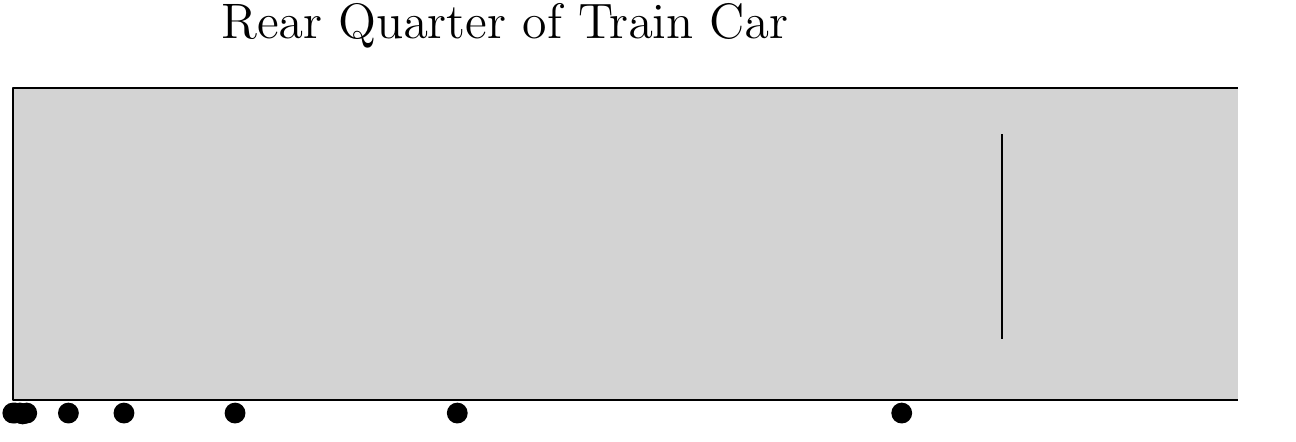}
\end{center}

\vspace{.5 cm}

\subsection{A Randomized Algorithm for Building Track}
Our algorithm is a simple example of the probabilistic method with alterations. In particular, we begin by randomly constructing a track that uses only a small number of pillars, and then we show that this track can be slightly altered in order to support the rear of the train car at every point.

We begin by installing each pillar randomly with probability $\frac{\ln n}{n}$. Even though this strategy has nothing to do with the structure of where the wheels are on the train, it already does remarkably well. In particular, if we place the train at some given position on the track, then there are $n$ different pillar positions that could potentially hold up the rear quarter. Each of these pillars positions has a $\frac{\ln n}{n}$ probability of having a pillar installed. It follows that, for a given position on the track, the rear quarter of the train has a
$$\left(1 -  \frac{\ln n}{n} \right)^{n}$$
probability of falling through the track. Taking advantage of the identity $\left(1 - \frac{1}{k}\right)^k \le \frac{1}{e}$, which holds for any $k \ge 1$, it follows that the wheels fall through the track with probability at most,
$$\frac{1}{e^{\ln n}} = \frac{1}{n}.$$

In other words, out of all the places we can place the train on the track, only a $\frac{1}{n}$-fraction of them will be problematic (in expectation). To fix this, we can just install one additional pillar to remedy each of these problematic positions. The result is a train track that fully supports the rear quarter of our train, and that uses only  $\left(\frac{1 + \ln n}{n}\right) \ell$ total feet of track, in expectation.

Of course, this isn't quite as good as we did when the wheels were evenly spaced out (we are a roughly $\ln n$ factor worse). But it's still pretty amazing! No matter how the wheels are distributed across each quarter of the train car, we can get away with installing only a $O\left(\frac{\ln n}{n} \right)$ fraction of the track!

The algorithm and analysis described above can be summarized in the following theorem:
\begin{thm}
  Consider a $4f$-foot long train car, and suppose that the rear
  quarter of a train car contains $n$ sets of wheels, each of which
  resides at a distinct integer distance from the rear of the car. In
  time $O(\ell n)$, one can construct an $\ell$-foot long train track
  with the following two properties: (1) as the train car drives down
  the track at least one set of wheels from the rear of the car is
  always touching track; and (2) the track consists of $1$-foot pillars,
  with the total expected number of pillars being at most
  $O\left(\frac{\ell \ln n}{n}\right)$.
\end{thm}

\section{A Matching Lower Bound}\label{sec:lower_bound}



In this section we show that using $O\left(\ell \frac{\ln n}{n} \right)$ pillars is optimal for some configurations of the train car. We are again going to use the probabilistic method, but this time in a more sophisticated way.

We continue to focus only on the rear quarter of the train car, which is $f$ feet long. We set $f = 2n$, and we construct the rear quarter of the train car by placing $n$ wheels at integer positions in the set $\{1, 2, \ldots, 2n\}$. We will then consider a track of length $\ell = 2f$, and show that $\Omega(\log n)$ pillars are necessary in order to support the rear quarter of the train car at all positions on the track. Recall that each pillar is one foot wide and is placed at an integer offset on the track.

Let $C$ be the set of wheel-positions in the rear quarter of the train car. We choose $C$ by placing a wheel at each position in $\{1, 2, \ldots, 2n\}$ independently with probability $\frac{1}{2}$. This means that $C$ has $n$ wheels in expectation, but may not actually have exactly $n$ wheels. The important thing to note is that, with at least $50\%$ probability, $C$ has $n$ or more wheels.

Now consider a track layout given by a subset $T$ of $\{1, 2, \ldots, 4n\}$, and satisfying $|T| = \frac{\ln n}{4}$. Whereas $C$ is a random variable, $T$ is a fixed set.

Define $X_{C, T}$ to be the event that, for every possible position of the rear quarter of the train on the track, at least one wheel from the rear quarter of the train is supported by track. From the perspective of the train car, $X_{C, T}$ is a good event. Formally, $X_{C, T}$ occurs if for every offset $k \in \{0, 1, \ldots, 2n\}$, we have $(k + C) \cap T \neq \emptyset$.\footnote{Recall that the rear quarter of the train car is length $2n$, and that the train track is length $\ell = 4n$. We only wish to consider offsets $k$ such that the rear quarter of the train car still sits entirely on potential track. That is, we wish to consider $k$ such that $(k + \{1, 2, \ldots, 2n\}) \subseteq \{1, 2, \ldots, 4n\}$, meaning that the values of $k$ that we care about are $k \in \{0, 1, \ldots, 2n\}$.}\footnote{For an integer $r$ and a set $S$, we use $r + S$ to denote the set $\{s + r \mid s \in S\}$.} 

The key to proving the desired lower bound is to show that the probability of $X_{C, T}$ occurring is very small, namely that $\Pr[X_{C, T}] < \frac{1}{2 \binom{4n}{(\ln n) / 4}}$. Because $T$ is a $\frac{\ln n}{4}$-element subset of $\{1, 2, \ldots, 4n\}$, there are at most $\binom{4n}{(\ln n) / 4}$ possibilities for $T$. Taking a union bound over all of these possibilities implies that
$$\Pr[X_{C, T} \text{ for any }T] < \frac{1}{2}.$$
On the other hand, we know that the number of wheels $|C|$ is less than $n$ with probability at most $1/2$. By a union bound, the probability that either $|C|$ has fewer than $n$ wheels or that $X_{C, T}$ holds for some $T$ is less than $1$. It follows that there must exist a car $C$ with $n$ or more wheels for which no track $T$ of size smaller than $(\ln n) / 4$ satisfies $X_{C, T}$. In fact, with slightly more careful bookkeeping, one can show that an even stronger property is true: almost all choices of how to place $n$ wheels in $C$ require a track of size larger than $(\ln n) / 4$ to support the car.

To complete the lower bound, the challenge becomes to show that $\Pr[X_{C, T}]$ is very small. That is, for a given choice of track $T$ containing $(\ln n)/4$ pillars, the probability that $T$ supports the rear-quarter of the train car $C$ is small.

Rather than examine the event $X_{C, T}$ directly, we instead examine a related quantity. Define $Y_{C, T}$ to be the number of positions $k \in \{0, 1, \ldots, 2n\}$ for which $(k + C) \cap T = \emptyset$ (i.e., the number of positions in which the rear quarter of the car, given by $C$, falls through the track $T$).

The relationship between $X_{C, T}$ and $Y_{C, T}$ is that $X_{C, T}$
occurs only if $Y_{C, T} = 0$. Our approach to completing the analysis
will be to first show that $\E[Y_{C, T}]$ is relatively large, and
then to show that the probability of $Y_{C, T}$ deviating
substantially from its expected value is small. This, in turn, implies
that $\Pr[Y_{C, T} = 0]$ is small, completing the analysis. In other
words, the problem of proving that there exists a train-car
configuration requiring a large amount of track is reduced to the
problem of proving a concentration inequality on the random variable
$Y_{C, T}$.

For each position $k \in \{0, 1, \ldots, 2n\}$, the set $T - k$ consists of $(\ln n) / 4$ elements. Since each of these elements is contained in $C$ with probability $1/2$, the probability that $C$ avoids all of the elements in $T - k$ is given by,
$$\frac{1}{2^{(\ln n) /4}} = \frac{1}{n^{1/4}}.$$
Summing over the values of $k$, it follows that the expected number of positions $k$ in which $(C + k) \cap T = \emptyset$ is
$$\E[Y_{C, T}] = 2n \cdot \frac{1}{n^{1/4}} > n^{3/4}.$$

The final step of the analysis is to prove a concentration inequality on $Y_{C, T}$. Standard Chernoff bounds do not apply here because $Y_{C, T}$ is not a sum of independent indicator random variables. Instead, we employ a more powerful tool, namely McDiarmid's Inequality:

\begin{thm}[McDiarmid '89 \cite{McDiarmid89}]
Let $A_1, \ldots, A_m$ be independent random variables over an arbitrary probability space. Let $F$ be a function mapping $(A_1, \ldots, A_m)$ to $\mathbb{R}$, and suppose $F$ satisfies,
$$\sup_{a_1, a_2, \ldots, a_m, \overline{a_i}} |F(a_1, a_2, \ldots, a_{i - 1}, a_i, a_{i + 1}, \ldots , a_m) - F(a_1, a_2, \ldots, a_{i - 1}, \overline{a_i}, a_{i + 1}, \ldots , a_m)| \le R,$$
for all $1 \le i \le m$. That is, if $A_1, A_2, \ldots, A_{i - 1}, A_{i + 1}, \ldots, A_m$ are fixed, then the value of $A_i$ can affect the value of $F(A_1, \ldots, A_m)$ by at most $R$; this is known as the \defn{Lipschitz Condition}. Then for all $S > 0$,
$$\Pr[|F(A_1, \ldots, A_m) - \E[F(A_1, \ldots, A_m)]| \ge R \cdot S] \le 2e^{-2S^2 / m}.$$
\end{thm}

To apply McDiarmid's Inequality to our situation, recall that $Y_{C, T}$ is defined to be the number of positions in the track $T$ that the rear quarter of the car, given by $C$, falls though. Whereas the track $T$ is fixed, each of the $2n$ possible wheels in $C$ is picked with probability $1/2$.
Define the indicator random variables $A_1, A_2, \ldots, A_{2n}$ so that $A_i$ indicates whether $i \in C$. As required by McDiarmid's Inequality, the $A_i$'s are independent of one-another, and $Y_{C, T}$ is a function of the $A_i$'s.

Now we show that the Lipschitz condition holds with $R = (\ln n) /
4$. Recall that the track $T$ consists of only $(\ln n) / 4$
pillars. Out of the $2n$ possible wheels $i$ that $C$ could contain,
each of those wheels $i$ is only relevant (to the car's stability)
when the car is $k$ feet down the track for some $k$ that places wheel
$i$ on top of a pillar. Since there are only $(\ln n)/4$ pillars, each
wheel $i$ is only relevant to the train car's stability for
$(\ln n) / 4$ positions $k$ on the track. In other words, for a given
wheel position $i \in \{1, 2, \ldots, 2n\}$, there are only
$(\ln n) / 4$ values of $k \in \{0, 1, \ldots, 2n\}$ for which
$(C + k) \cap T$ can possibly contain $i$.  As a result, each $A_i$
can only affect the value of $Y_{C, T}$ by at most $(\ln n) / 4$,
meaning that the Lipschitz condition holds with $R = (\ln n) / 4$.

Applying McDiarmid's Inequality, we get that
$$\Pr[n^{3/4} - Y_{C, T} >  n^{5/8} \cdot (\ln n) / 4] \le 2e^{-n^{1/4}}.$$
When $n$ is large, this probability is much smaller than $\frac{1}{2 \binom{4n}{(\ln n) / 4}}$. It follows that $\Pr[X_{C, T}] = \Pr[Y_{C, T} = 0] < \frac{1}{2 \binom{4n}{(\ln n) / 4}}$. Summing over all possible options for the track $T$, the probability that any of them support the train car $C$ is therefore less than $1/2$. It follows that some train car $C$ with $n$ or more wheels fails to be supported by any track $T$ consisting of $(\ln n)/4$ or fewer pillars. This completes the lower bound, and establishes the following theorem.

\begin{thm}
  There exists a set of wheel positions
  $C \subseteq \{1, 2, \ldots, 2n\}$ such that $|C| \ge n$, and such
  that in order for a track $T \subseteq \{1, 2, \ldots, 4n\}$ to
  support the set of wheels at every position (meaning that
  $(C + k) \cap T \neq \emptyset$ for each $k \in \{0, \ldots, 2n\}$)
  the track $T$ must have size $\Omega\left(\frac{\ln n}{n}\right)$.
\end{thm}

\section{Three Algorithms for Building Track}\label{sec:three_algorithms}

In this section, we revisit the problem of constructing a train track
that uses $O\left(\frac{\ell \ln n}{n}\right)$ feet of track, and present three
alternative algorithms for this problem, each of which gives the same
guarantees as the algorithm in Section \ref{sec:upper_bound}.

We continue to assume that the wheels of the train car are at
integer positions, and we focus only on the $n$ wheels in the rear
quarter of the train car. We use $C$ to denote the set of positions of
wheels, meaning that $C$ is an $n$-element subset of
$\{1, \ldots, f\}$. Our goal is to construct a set of pillars
$T \subseteq \{1, 2, \ldots, \ell\}$ such that for each
$k \in \{0, 1, \ldots, \ell - f\}$, the set $(C + k) \cap T$ is
non-empty. As was the case in Section \ref{sec:upper_bound}, we want an algorithm
that runs in expected time $O(n \ell)$ and produces a set $T$ with
expected size $O\left(\frac{\ell \ln n}{n} \right)$.

Each of the three algorithms applies a different core
technique from the overlap of probabilistic combinatorics and
randomized algorithms:
\begin{itemize}
\item \textbf{A Deterministic Algorithm (Section
    \ref{sec:deterministic}).} The first algorithm uses the method of
  conditional probabilities to derandomize the algorithm given in
  Section \ref{sec:upper_bound}. 
\item \textbf{An Application of the Algorithmic Lov\'asz Local Lemma
    (Section \ref{sec:Lovasz})} The second algorithm uses the
  algorithmic version of the Lov\'asz Local Lemma due to Moser and
  Tardos \cite{MoserTa10}.
\item \textbf{An Application of the Min-Hash Technique (Section
    \ref{sec:minhash})} The final algorithm uses a variant of the
  min-hash technique, which has previously found important
  applications in locality sensitive hashing and string alignment \cite{Broder97, BroderCh00, Manasse12, Kuszmaul19, CharikarGeKi18, OndovTe16}.
\end{itemize}

\subsection{A Deterministic Algorithm}\label{sec:deterministic}

In this section, we use the method of conditional probabilities \cite{AlonSp04} in order to design a deterministic algorithm for the train-track problem.

The basic idea behind the method of conditional probabilities is as follows. Suppose $X_1, \ldots, X_\ell$ are independent real-valued random variables, and that $F: \mathbb{R}^\ell \rightarrow \mathbb{R}$ is an objective function that we wish to minimize. We are given that $\E[F(X_1, \ldots, X_\ell)] \le R$ for some value $R$, and we wish to find values of $x_1, \ldots, x_\ell \in \mathbb{R}$ for which $F(x_1, \ldots, x_{\ell}) \le R$. The method of conditional probabilities takes an iterative approach. Suppose we already have values of $x_1, \ldots, x_k$ such that
$$\E[F(x_1, \ldots, x_k, X_{k + 1}, \ldots, X_\ell)] \le R.$$
Then there must exist some value $x_{k + 1}$ such that
\begin{equation}
  \E[F(x_1, \ldots, x_k, x_{k + 1}, X_{k + 2}, \ldots, X_\ell)] \le R.
  \label{eq:iterative_fill}
\end{equation}
The key challenge in applying the method of conditional probabilities is to design an objective function $F$ that both captures the problem at hand, but that also allows for one to efficiently determine which value of $x_{k + 1}$ satisfies \eqref{eq:iterative_fill}. This, in turn, allows for one to iteratively determine values for all of $x_1, \ldots, x_\ell$ such that $F(x_1, \ldots, x_\ell) \le R$.

In order to apply the method of conditional probabilities to the train-track problem, we define $X_1, \ldots, X_\ell$ to be zero-one random variables, each of which takes value $1$ with probability $\frac{\ln n}{n}$. Given values $x_1, \ldots, x_\ell$ for random variables $X_1, \ldots, X_\ell$, we can construct a train track $T$ by first setting $T_1 = \{i \mid x_i = 1\}$, and then defining $T$ to be $T_1$ with one additional pillar for each position $k$ in which the train wheels $C$ fall through the track $T_1$. Since our goal is to minimize the size of $T$, we define our objective function to be $F(x_1, \ldots, x_\ell) = |T|$.

In Section \ref{sec:upper_bound}, we showed that $\E[F(X_1, \ldots, X_\ell)] \le (1 + \ln n) / n$. Suppose that we have values $x_1, \ldots, x_k \in \{0, 1\}$ such that
\begin{equation}
  \E[F(x_1, \ldots, x_k, X_{k + 1}, \ldots, X_\ell)] \le (1 + \ln n) / n.
  \label{eq:xk}
\end{equation}
Moreover, suppose that we maintain values $p_0, \ldots, p_{\ell - f}$ such that each $p_i$ denotes the probability that the set of wheels $(C + i)$ fall through the track $T = \{j \mid x_j = 1\} \cup \{j \mid X_j = 1\}$. This means that we can compute $\E[F(x_1, \ldots, x_k, X_{k + 1}, \ldots, X_\ell)]$ as
\begin{equation}
  \Big|\{i \mid x_i = 1\}\Big| + \frac{\ln n}{n} (\ell - k) + \sum_i p_i.
  \label{eq:evaluating_F}
\end{equation}
The first two terms represent $\E[|T_1|]$, and the third term represents $\E[|T \setminus T_1|]$. 

Given values of $x_1, \ldots, x_k$ such that \eqref{eq:xk} holds, we wish to find a value of $x_{k + 1} \in \{0, 1\}$ so that \eqref{eq:xk} will hold for $k + 1$. If we set $x_{k + 1} = 1$, then this has the effect of increasing the expected initial size of $|T_1|$ by $1 - \frac{\ln n}{n}$, and of zeroing out any $p_i$'s for which $k + 1 \in (C + i)$. On the other hand, if we set $x_{k + 1} = 0$,  then this has the effect of decreasing the expected initial size of $|T_1|$ by $\frac{\ln n}{n}$ and replacing each $p_i$ for which $k + 1 \in (C + i)$ with $\frac{p_i}{1 - (\ln n) / n}$. It follows that in time $O(n)$, one can update \eqref{eq:evaluating_F} in order to determine $\E[F(x_1, \ldots, x_{k + 1}, X_{k + 2}, \ldots, X_\ell)]$ for each of the two possible values of $x_{k + 1}$. By selecting the value that minimizes the expected objective function, we can guarantee that
$$\E[F(x_1, \ldots, x_{k + 1}, X_{k + 2}, \ldots, X_\ell)] \le \frac{1 + \ln n}{n}.$$
Continuing like this, we can find values of $x_1, \ldots, x_\ell$ such that $F(x_1, \ldots, x_\ell) \le (1 + \ln n) / n$ in time $O(\ell n)$. Using these $x_1, \ldots, x_\ell$ to construct the track $T$ results in a track that uses at most $(1 + \ln n) / n$ pillars, as desired.

\subsection{An Application of the Algorithmic Lov\'asz Local Lemma}\label{sec:Lovasz}

Given a large collection of unlikely events $E_1, \ldots, E_m$, such
that each event $E_i$ is related to only a small number of other
events $E_j$, the Lov\'asz Local Lemma is a technique for showing that
there exists a way for all $m$ events to mutually not occur. In one of its
most basic forms, the Lov\'asz Local Lemma can be stated as follows:

\begin{thm}[Lov\'asz and Erd\"os '73 \cite{LovaszEr73}]
Suppose $X_1, \ldots, X_s$ are independent random variables, possibly over different probability spaces. Let $E_1, \ldots, E_m$ be events such that each $E_i$ is determined by some subset of the $X_j$'s -- that is, there exists an index set $I_i \subseteq [s]$ such that $E_i$ is determined by the outcome of the $X_j$'s for which $j \in I_i$. Say that two events $E_i$ and $E_j$ \defn{depend on each other} if $I_i \cap I_j \neq \emptyset$. Let $p$ be such that $\Pr[E_i] \le p$ for each $i$, and let $d$ be such that each $E_i$ depends on at most $d$ different $E_j$'s (including $E_i$). If $p  d  e \le 1$, where $e$ is the universal constant, then there is a positive probability that none of the events $E_1, \ldots, E_m$ occur.
\label{lem:lovasz_nonalg}
\end{thm}

The algorithmic version of the Lov\'asz Local Lemma gives an efficient algorithm for constructing values of $X_1, \ldots, X_s$ in order to ensure that none of the events $E_1, \ldots, E_m$ occur.
\begin{thm}[Moser and Tardos '10 \cite{MoserTa10}]
  Suppose that the conditions from Theorem \ref{lem:lovasz_nonalg}
  hold. Consider the following algorithm for choosing values of
  $X_1, \ldots, X_s$: First independently sample each of
  $X_1, \ldots, X_s$ from its defining probability distribution; then,
  as long as their exists at least one event $E_i$ that holds, pick
  such an event $E_i$ and resample the $X_j$'s for each $j \in
  I_i$. Each time that the $X_j$'s are resampled for some event $E_i$,
  we call the resamplings a \defn{phase} of the algorithm. The
  algorithm terminates once the $X_i$'s have been assigned values that
  result in no events $E_i$ occurring.

  The above algorithm, known as the \defn{fix-it algorithm},
  terminates in finite expected time, and the expected number of
  phases is at most $n / d$.
\end{thm}

In order to apply the Algorithmic Lov\'asz Local Lemma to our problem,
we define $X_1, \ldots, X_\ell$ to be independent zero-one random
variables, each taking value $1$ with probability $\frac{1 + 2\ln n}{n}$.
Each $X_i$ is the indicator variable for whether we include pillar $i$
in the train track. We then define events $E_0, \ldots, E_{\ell - f}$
so that $E_i$ is the event that the rear-quarter of the train car
falls through the track at position $i$. That is, $E_i$ occurs if
$(C + i) \cap \{j \mid X_j = 1\} = \emptyset$.

Each event $E_i$ depends on only $n$ different $X_j$'s, and each $X_j$ is relevant to only $n$ different $E_k$'s. It follows that each event $E_i$ depends on at most $n^2$ other $E_k$'s (including $E_i$). This means that we can apply the Algorithmic Lov\'asz Local Lemma with $d = n^2$.

In order for a given event $E_i$ to occur, there are $n$ different pillars that all must fail to appear in the track. The probability of this happening is
$$\Pr[E_i] = \left(1 - \frac{1 + 2\ln n}{n}\right)^n < \frac{1}{e^{1 + 2 \ln n}} \le \frac{1}{en^2}.$$

Using $d = n^2$ an $p = \frac{1}{en^2}$, we can apply the Lov\'asz
Local Lemma in order to conclude that there exists a choice of pillars
$X_1, \ldots, X_\ell$ so that the wheels in $C$ are supported along the
entire track. This alone is not a useful observation since, of course,
setting $X_1, \ldots, X_\ell$ all to $1$ would trivially support the
wheels in $C$ at all points. On the other hand, if we apply the
algorithmic version of the Lov\'asz Local Lemma, then get an
additional fact: that the fix-it algorithm terminates after only
$O(\ell / n^2)$ phases in expectation.

Since each phase resamples only $n$ different $X_i$'s, the resamples
contribute at most $O(\ell / n)$ pillars in expectation. On the other
hand, the initial configuration of the $X_i$'s contributes at most
$(1 + 2\ln n) / n$ pillars in expectation. It follows that, at the end
of the fix-it algorithm, the resulting track configuration will use
at most $(2 + 2 \ln n) / n$ pillars in expectation. A careful
implementation of the fix-it algorithm will run in expected time
$O(\ell n)$, as desired.

\subsection{An Application of Min-Hash}\label{sec:minhash}

Given a collection of sets $\mathcal{S}$, \emph{Min-Hashing} is a technique for randomly sampling one element for each set $S \in \mathcal{S}$. The technique works by first hashing each element $s$ of each set $S$ in $\mathcal{S}$ to a random real number $h(s) \in (0, 1)$. For each set $S \in \mathcal{S}$, one then samples the element $s \in S$ with minimum hash $h(s)$.

The Min-Hashing technique plays important roles in both Locality Sensitive Hashing \cite{Broder97, BroderCh00, Manasse12} and string-alignment algorithms \cite{Kuszmaul19, CharikarGeKi18, OndovTe16}. The key property of Min-Hashing is that if two sets $S_1, S_2 \in \mathcal{S}$ are similar to one-another, then their min-hash is likely to be the same. And more generally, if an element $s$ is the minimum-hashed element in one set $S \in \mathcal{S}$, then $s$ is likely to also be the minimum-hashed element in other sets.

In our application of Min-Hashing, we need not actually use hash functions. Instead, we assign random real numbers $r_1, \ldots, r_\ell \in (0, 1)$ to each of the $\ell$ possible track pillars. For each possible offset $k \in \{0, 1, 2, \ldots, \ell - f\}$, define the set $S_k = (C + k)$ to be the positions that the wheels in $C$ take when the train car is $k$ feet down the track. We construct a train track $T$ by adding the pillar $\operatorname{argmin}_{s \in S_k} r_s$ for each set $S_k$. That is, for each position that the train could sit in the track, we look at all possible pillars that could hold the rear-quarter of the train up, and we include in our track the pillar with the minimum assigned random value $r_s$. We say that this pillar $s$ is \defn{sampled} from $S_k$. 

By construction, the set of pillars $T$ is guaranteed to support the wheels $C$ at every position. What is less clear is whether $|T|$ will be small. Here is where we take advantage of the properties of Min-Hashing, and the fact that many of the sets $S_k$ sample the same pillars as one another.

The key observation is that almost all of the pillars $s$ that are sampled have small random values $r_s$. Consider, in particular, the probability that for a given set $S_k$, we sample a pillar $s$ for which $r_s > (\ln n) / n$. This means that all $n$ pillars in $S_k$ were assigned random values larger than $(\ln n) / n$, which happens with probability at most,
$$\left(1 - \frac{\ln n}{n}\right)^n \le \frac{1}{e^{\ln n}} = \frac{1}{n}.$$

It follows that, out of the $\ell - f$ samplings that occur, the expected number of pillars $s$ for which $r_s > (\ln n) / n$ that are sampled is at most $(\ell - f) / n \le \ell / n$. On the other hand, even if every pillar $s$ for which $r_s \le (\ln n) / n$ is sampled, the expected number of them is at most $\ell (\ln n) / n$. The total number of sampled pillars, and thus the size of $T$, is therefore at most $\ell (1 + \ln n) / n$, in expectation. This completes the analysis of the algorithm.


\bibliographystyle{plainurl} \bibliography{train}

\begin{thebibliography}{10}

\bibitem{AlonSp04}
N.~Alon and J.~H. Spencer.
\newblock {\em The probabilistic method}.
\newblock John Wiley \& Sons, 2004.

\bibitem{Broder97}
A.~Z. Broder.
\newblock On the resemblance and containment of documents.
\newblock In {\em Proceedings. Compression and Complexity of Sequences 1997
  (Cat. No. 97TB100171)}, pages 21--29. IEEE, 1997.

\bibitem{BroderCh00}
A.~Z. Broder, M.~Charikar, A.~M. Frieze, and M.~Mitzenmacher.
\newblock Min-wise independent permutations.
\newblock {\em Journal of Computer and System Sciences}, 60(3):630--659, 2000.

\bibitem{CharikarGeKi18}
M.~Charikar, O.~Geri, M.~P. Kim, and W.~Kuszmaul.
\newblock On estimating edit distance: Alignment, dimension reduction, and
  embeddings.
\newblock In {\em 45th International Colloquium on Automata, Languages, and
  Programming (ICALP 2018)}. Schloss Dagstuhl-Leibniz-Zentrum fuer Informatik,
  2018.

\bibitem{LovaszEr73}
P.~Erd{\H{o}}s and L.~Lov{\'a}sz.
\newblock Problems and results on 3-chromatic hypergraphs and some related
  questions.
\newblock In {\em Colloquia Mathematics Societatis Janos Bolai 10. Infinite and
  Finite Sets, Keszthely (Hungary)}. Citeseer, 1973.

\bibitem{Kuszmaul19}
W.~Kuszmaul.
\newblock Efficiently approximating edit distance between pseudorandom strings.
\newblock In {\em Proceedings of the Thirtieth Annual ACM-SIAM Symposium on
  Discrete Algorithms}, pages 1165--1180. Society for Industrial and Applied
  Mathematics, 2019.

\bibitem{Manasse12}
M.~S. Manasse.
\newblock On the efficient determination of most near neighbors: horseshoes,
  hand grenades, web search and other situations when close is close enough.
\newblock {\em Synthesis Lectures on Information Concepts, Retrieval, and
  Services}, 4(4):1--88, 2012.

\bibitem{McDiarmid89}
C.~McDiarmid.
\newblock On the method of bounded differences.
\newblock {\em Surveys in combinatorics}, 141(1):148--188, 1989.

\bibitem{MoserTa10}
R.~A. Moser and G.~Tardos.
\newblock A constructive proof of the general lov{\'a}sz local lemma.
\newblock {\em Journal of the ACM (JACM)}, 57(2):11, 2010.

\bibitem{OndovTe16}
B.~D. Ondov, T.~J. Treangen, P.~Melsted, A.~B. Mallonee, N.~H. Bergman,
  S.~Koren, and A.~M. Phillippy.
\newblock Mash: fast genome and metagenome distance estimation using minhash.
\newblock {\em Genome biology}, 17(1):132, 2016.

\end{thebibliography}
\end{document}